\documentclass[conference]{IEEEtran}
\IEEEoverridecommandlockouts
\usepackage[natbib=true,
            style=numeric,
            sorting=none,
            backend=biber,
            ]{biblatex}
\usepackage{amsmath,amssymb,amsfonts}
\usepackage[noend]{algpseudocode}
\usepackage{algorithm}
\algnewcommand{\LineComment}[1]{\State \(\triangleright\) #1}
\usepackage{graphicx}
\usepackage{textcomp}
\usepackage{xcolor}
\usepackage[utf8]{inputenc}
\usepackage[T1]{fontenc}
\usepackage{cleveref}
\usepackage{tabularx}
\usepackage{multirow}
\usepackage{balance}
\usepackage{wrapfig}
\usepackage[font=normalsize, margin=0mm]{caption}
\usepackage[font=small]{subfig}

\title{\LARGE \bf
Network Localization Based Planning for Autonomous Underwater Vehicles with Inter-Vehicle Ranging
}
\author{
\IEEEauthorblockN{Alan Papalia}
\IEEEauthorblockA{\textit{Computer Science and Artificial Intelligence Laboratory} \\
\textit{Massachusetts Institute of Technology}\\
Cambridge, MA, USA \\
apapalia@mit.edu}
\and
\IEEEauthorblockN{John Leonard}
\IEEEauthorblockA{\textit{Computer Science and Artificial Intelligence Laboratory} \\
\textit{Massachusetts Institute of Technology}\\
Cambridge, MA, USA \\
jleonard@mit.edu}
}

\newcommand{\comment}[1]{}

\begin{document}

\maketitle
\thispagestyle{empty}
\pagestyle{empty}
\begin{abstract}
 Localization between a swarm of AUVs can be entirely estimated through the use of range measurements
 between neighboring AUVs via a class of techniques commonly referred to as sensor network localization.
 However, the localization quality depends on network
 topology, with degenerate topologies, referred to as low-rigidity
 configurations, leading to ambiguous or highly uncertain localization results. This
 paper presents tools for rigidity-based analysis, planning, and control
 of a multi-AUV network which account for sensor noise and limited sensing
 range. We evaluate our long-term planning framework in several two-dimensional
 simulated environments and show we are able to generate paths in feasible time
 and guarantee a minimum network rigidity over the full course of the paths.
\end{abstract}

\section{INTRODUCTION}

Multi-AUV swarms bear the promises of increased coverage, greater efficiency in
ocean deployments, and improved ability to track spatially and temporally
dynamic oceanographic phenomena. However, the lack of Global Navigation
Satellite System (GNSS) and sparsity of salient features for navigation present
substantial challenges in accurately localizing each AUV underwater, a required
ability for real-world deployment. Additionally, severe communication
constraints due to the underwater environment limit multi-AUV systems to
techniques which require minimal communications. These combined challenges in
localization and communication greatly limit the deployment of multi-AUV
networks.

The simultaneous localization and mapping community has developed a large number
of techniques which use observations of environmental features to allow for
autonomous localization, but these techniques often extend poorly to the
underwater domain due to a common sparsity of features to observe, large amount
of error in observations made, and oftentimes a need for relatively large data
transmissions \cite{Paull2014}. Previous works \cite{Larsen2000} in single-AUV
localization rely upon the use of costly inertial navigation systems which offer
greatly reduced drift in dead-reckoned estimates. However, such systems can
often cost hundreds of thousands of dollars, which becomes cost-prohibitive when
scaling to multi-AUV swarms. Recent work \cite{Rypkema2018} proposed one-way
acoustic communications to perform formation control in which `followers' keep
relative positioning to a `leader'. While this allows for localization and
reduced communication, this limits configurations to predetermined geometries
and relies on functionality of the `leader'.

Sensor network localization \textbf{(SNL)} is a promising set of localization
techniques for multi-AUV networks which use inter-AUV range measurements to
estimate relative position of network members \cite{Aspnes2006}. Furthermore,
knowing the absolute location of four network members, via GNSS or inertial
navigation, is fully sufficient to fix the absolute coordinates of the entire
network and provide absolute three dimensional localization of all others. This
allows for complete localization using only range measurements and depending on
the operational requirements can enable localization in which a small number of
AUVs use GNSS fixes to maintain absolute localization for the entire network.
Such an approach allows use of recently developed low-cost ranging technology
\cite{Fischell2020} and absolves need for high-cost inertial navigation solutions or fixed acoustic beacons.

\begin{figure}[!htbp]
  \centering
  \includegraphics[height=4.75cm, width=\linewidth ]{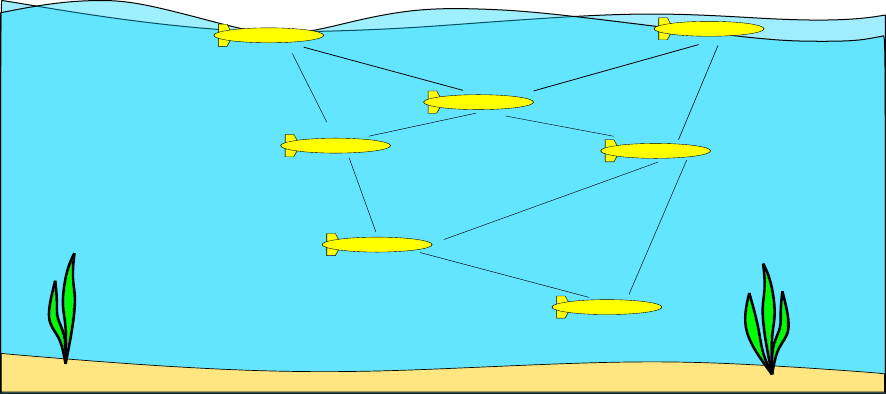}
  \label{fig:auv-network}
  \caption{Example multi-AUV network. Edges between AUVs represent
    distance measurements between corresponding vehicles. Surfacing AUVs
    can use GNSS for absolute positioning, thus
    serving as reference positions for the entire swarm and allowing
    absolute localization of the entire swarm. }
\end{figure}

Many distributed and centralized SNL techniques have been developed to robustly
perform localization in the presence of noisy range measurements
\cite{Calafiore2012,Biswas2006}. However, it is well known that the accuracy of
SNL approaches depend on the topology of the network, with certain 'poor' network configurations causing high localization error \cite{Moore2004}.

We consider the problem of generating paths to a set of goal locations for a
multi-AUV network while ensuring that the network remains in configurations
robust to noise. We use the \emph{rigidity eigenvalue} \cite{LeNy2018} to
measure the robustness of a configuration, which we term \emph{network
rigidity}. Previous work considering similar networks of robots
\cite{Zelazo2015} developed a linear algebraic representation of network
rigidity and proposed gradient-based controls to maintain network rigidity. This
work was extended by \cite{LeNy2018} in which the linear-algebraic
representation of network rigidity was modified to consider an information
theoretic interpretation of network rigidity and the use of potential-field
planning was proposed. However, both of these approaches relied on the use of
gradient-based controls to maneuver such multi-agent networks, which can cause
highly inefficient paths to be generated and are inherently at risk of getting
stuck in local-minima during planning and therefore prevented from reaching
target locations.

To address these issues we formulate a prioritized path-planning framework
\cite{VanDenBerg2005} which leverages the underlying structure of network
rigidity to more efficiently compute paths which enforce minimum network
rigidity. To make our algorithm more robust to the possibility of a valid set of
trajectories not being found, we make use of a conflict identifying mechanism
which attempts to find trajectories which degrade the rigidity of the network.
Though our approach differs from previous works in conflict-based multi-agent
planning \cite{sharon2015conflict}, there is similarity in that conflicts are
used to aid planning. We evaluate our approach in simulation and show our
framework successfully finds paths in reasonable time while maintaining minimum
network rigidity.
\section{METHODS}

We seek to develop a framework for scalable, rigidity constrained multi-AUV
path-planning that allows for unambiguous and robust localization via SNL
techniques. For simplicity, we constrain ourselves to a two-dimensional system,
but note that extension to higher dimensions is straightforward.

\subsection{Network Rigidity}

We quantify network rigidity based on the eigenvalues of a network's derived
Fisher information matrix (FIM). As in \cite{LeNy2018, patwari2005locating}, for
a given network configuration under the assumption of Gaussian measurement noise
we define the Fisher information matrix, $F=A^TA$ where $A \in \mathbb{R}^{m
    \times dn}$, $m$ is the number of measurements, $d$ the dimensionality of the
system, and $n$ the number of nodes. The rows of $A$ correspond to measurements
between network nodes (AUVs), with a single row per measurement. Defined below,
$row_{m}$ denotes the row corresponding to measurement $m$ with standard
deviation $\sigma$ between nodes $i$ and $j$. The value of $\alpha$ depends on
whether the measurement model represents Gaussian additive or multiplicative
noise.

It it proven that the inverse of the FIM is the lower
bound on the variance of an unbiased estimator, commonly referred to as the
Cram\'er-Rao Lower Bound (CRLB) \cite{barfoot2017state}. Though the actual
estimation variance relies on the estimation technique used and the CRLB cannot
be achieved in many cases, this provides an
information-theoretic limit for the uncertainty of an estimation technique.
Importantly, this implies that in our specific case the eigenvalues of the FIM
have an inverse relationship with the best possible localization uncertainty.
That is, lower eigenvalues correspond to increased best-case uncertainty in
localization results.

\begin{IEEEeqnarray} {rCl}
  \label{eqn:fim-eqns}
  &\Delta_x &= x_i - x_j \\
  &\Delta_y &= y_i - y_j \\
  &L &= \sqrt{\Delta_x^2 + \Delta_y^2}  \\
  \nonumber   \\
  &\alpha &=
  \begin{cases}
    1 & \textit{Additive Noise}       \\
    2 & \textit{Multiplicative Noise} \\
  \end{cases} \\
  \nonumber   \\
  & row_{m} &=  \dfrac{1}{\sigma L^{\alpha}}
  \begin{cases}
    \Delta_x  & \textit{index}=2i   \\
    \Delta_y  & \textit{index}=2i+1 \\
    -\Delta_x & \textit{index}=2j   \\
    -\Delta_y & \textit{index}=2j+1 \\
    0         & \textit{otherwise}
  \end{cases} \\
  \nonumber
\end{IEEEeqnarray} \

Because of the natural relationship between the eigenvalues of the FIM and the
localization uncertainty, we use these eigenvalues as a heuristic measure for
the localizability of a network configuration, or, as was previously introduced,
the \emph{network rigidity}. Similar to \cite{LeNy2018,Zelazo2015}, we define
network rigidity for a given sensor network as the least nontrivial eigenvalue
of $F$, where trivial eigenvalues are all zero and have been shown
\cite{Zelazo2015} to correspond to the degrees of freedom (DOF) of a the special
Euclidean space the network occupies (e.g. two-dimensional space being 3-DOF).
It has been theoretically shown that a non-zero network rigidity is required to
find unique solutions to the range-only localization problem \cite{Aspnes2006}.

To heuristically control the localizability of a network we will enforce a
minimum rigidity during the trajectory planning sequence, below which a network
will be considered nonrigid and disallowed. This causes our proposed planning
technique to perform a large number of eigenvalue computations. For this reason,
it is important to note that it directly follows from the construction of
$F=A^TA$ that $F$ is a positive semidefinite matrix, and thus has only real,
non-negative eigenvalues. In practice we leverage this fact to use computational
methods that are specially designed for Hermitian matrices, which positive
semidefinite matrices are a subclass of, to reduce the time required to perform
these eigenvalue computations.

\subsection{Path-Planning}

\comment{TODO better picture of planning graph}

From here our approach takes the form of priority-based multi-agent planning on
a single graph. In our methodology planning on a graph is meant to indicate that
all agents share a common graph in which nodes are locations, edges connect
neighboring locations according to some set of predetermined rules, that the
agents are only considered to occupy the nodes of the graph, and that the agents
move through the world by moving along these edges between neighboring nodes
\cite{Kavraki1996}. Priority-based planning indicates that each agent
individually performs planning on this graph in a predetermined sequence.

In the planning process we impose the constraint that no two agents can occupy
the same location at the same time and that network rigidity must remain above a
pre-specified minimum rigidity for all timesteps. In addition, we apply a
heuristically driven technique which attempts to avoid low-rigidity
configurations early in the planning sequence. This is to prevent going through
the entire planning sequence when early in the planning sequence the already
planned trajectories would likely result in low-rigidity configurations
regardless of the subsequent trajectories

Though network rigidity is not guaranteed to monotonically increase as agents
are added to the network, as a heuristic measure to avoid low-rigidity
configurations early in the planning sequence we enforce that every agent must
plan a path which stays entirely within the valid set, $V$, defined later in
this section. This reduces the search space, reducing necessary computation at
the cost of potentially rejecting valid full-network trajectories.

This framework was chosen to address several practical concerns in multi-agent
planning. We look to avoid gradient-based methods due the previously mentioned
issues of local minima, particularly as the number of AUVs increase and the
potential fields used run the risk of growing complexity and increased number of
local minima. Similarly, as the number of AUVs increase the dimensionality of
the planning space increases. The use of priority planning reduces the dimension
of the planning space from $(n \times d)$ to $d$.

Furthermore, planning on graphs allows the use of sampling based planning
techniques, which have empirically found success in reducing the combinatorial
cost of high-dimensional problems such as multi-agent planning. Importantly, the
graph-based structure also allows for straightforward bookkeeping and recycling
of certain computations as the planning process occurs; this is discussed in the
remainder of this section.

To perform this bookkeeping and reduce unnecessary computation during planning
we use multiple set-based representations of locations on the planning-graph,
with there being distinct collections of these sets that correspond to each AUV
and timestep. We will first describe the sets and how they are constructed and
then explain how they are used to perform bookkeeping and eliminate unnecessary
computation. As our planning technique is priority-based, sets for AUV $0$ are
quickly computed upon construction of the planning graph. For all other AUVs,
the sets for AUV $n$ are computed after planning by AUV $n-1$.

For the $ith$ AUV at time $t$ we denote the sets: reachable
($\mathbf{P_{i,t}}$), connected ($\mathbf{C_{i,t}}$), rigid
($\mathbf{R_{i,t}}$), and valid ($\mathbf{V_{i,t}}$). $\mathbf{C_{i,t}}$ is the
set of all states which would be within sensing radius of AUVs $(0, 1, ...,
  i-1)$ at time $t$. $\mathbf{R_{i,t}}$ is the set of all states which AUV $i$
could occupy at time $t$ that would allow for AUVs $0,1,..,i$ to form a network
that satisfies the minimum rigidity constraint. $\mathbf{P_{i,t}}$ is the set of
all locations which are connected to a location in $\mathbf{V_{i,t-1}}$ for AUV
$i$ at time $t$. $\mathbf{V_{i,t}}$ is the set of allowable states, which the
AUVs are required to use for planning. We define $\mathbf{V_{i,t}}$ and note the
relationship between $\mathbf{R_{i,t}}$ and $\mathbf{C_{i,t}}$ in Equations
(\ref{eqn:rigid-sets}$-$\ref{eqn:valid-sets}).

The purpose of these sets of graph locations is to track which
locations AUV $i$ can inhabit at time $t$ while remaining part of a rigid
network without having to exhaustively check every single location. As we know
that an AUV in \(d\)-dimensional space must have \(d\) range measurements from other
AUVs to be rigidly connected, we can avoid checking for network rigidity until
we know an AUV would be within sensing range of \(d\) AUVs at that given time and
location.This is the purpose of the connected set, \(\mathbf{C_{i,t}}\), to
ensure that a location is connected before it is checked for rigidity.
Similarly, there is no need to check for rigidity if a location is not
reachable, that is to say there is no path an AUV can follow from timestep \(0\)
to timestep \(t\) to get to that location while staying entirely within the valid
sets for that AUV at all previous timesteps.

\begin{IEEEeqnarray} {rCl}
  & \mathbf{R_{i,t}} \subseteq  \mathbf{C_{i,t}}  \label{eqn:rigid-sets}\\
  & \mathbf{V_{i,t}} \equiv
  \begin{cases}
    \mathbf{P_{i,t}}                       & i=0       \\
    \mathbf{P_{i,t}} \cap \mathbf{C_{i,t}} & i=1       \\
    \mathbf{P_{i,t}} \cap \mathbf{R_{i,t}} & otherwise \\
  \end{cases} \label{eqn:valid-sets}
\end{IEEEeqnarray}

\begin{figure}[!htbp]
  \centering
  \includegraphics[width=.7\linewidth, height=.35\textheight]{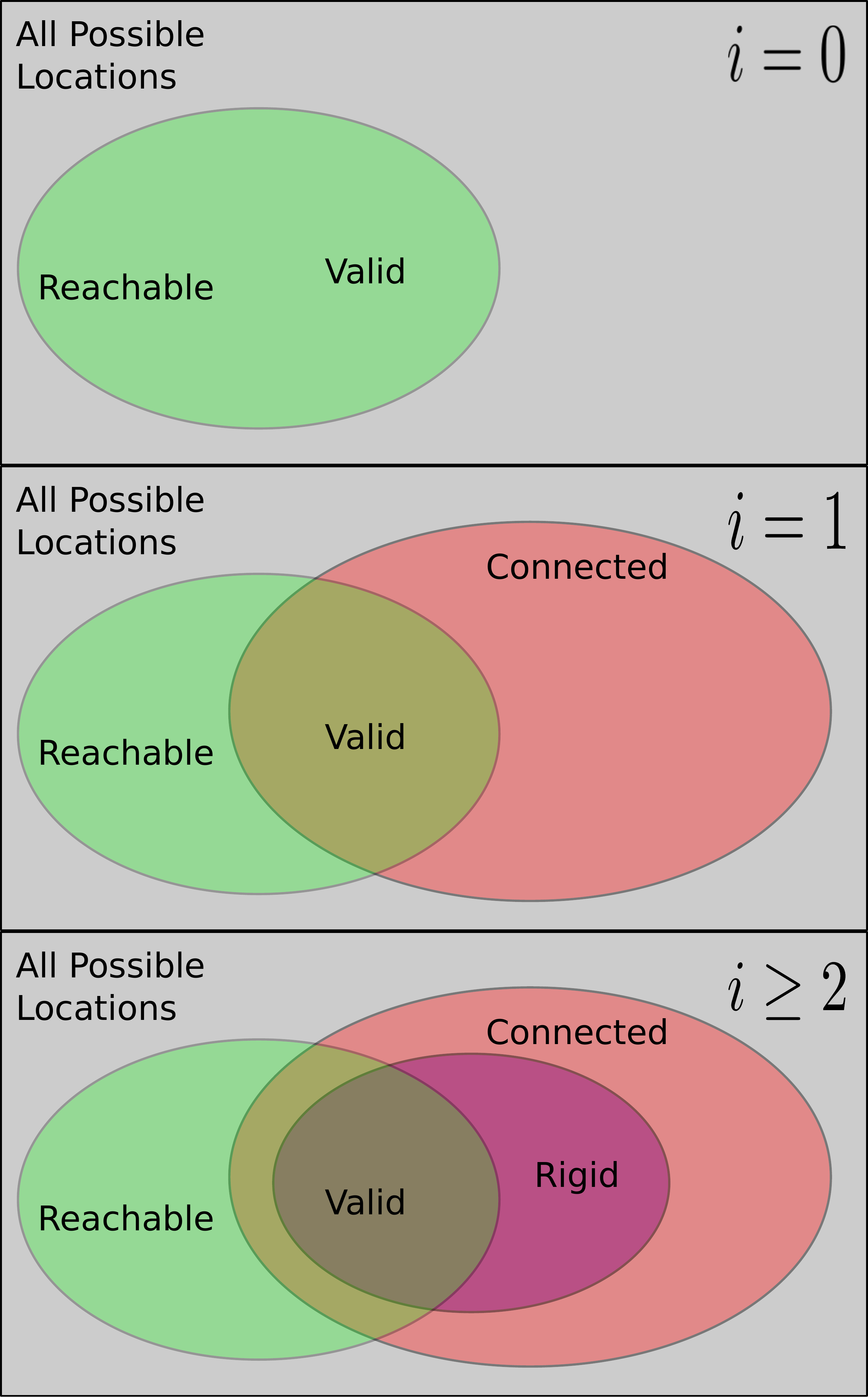}
  \caption{Visual representation of relationships between planning sets. \textbf{(Top)}
    the planning sets for AUV \(0\), where all reachable locations are considered
    valid. \textbf{(Middle)} the planning sets for AUV \(1\), where connectivity to
    AUV \(0\) is enforced. \textbf{(Bottom)} the planning sets for all AUVs \(i \geq
    2\) where rigidity is enforced at every planning step.}
  \label{fig:planning-sets}
\end{figure}

\begin{algorithm}
  \caption{Perform rigidity-constrained priority planning for all AUVs in a sequence} \label{alg:planning}
  \begin{algorithmic}[1]
    \Procedure{Multi-AUV Planning}{}
    \State \(\text{trajectories} \gets \emptyset \)
    \State \(\text{conflicts} \gets \emptyset \)
    \State \(V_0 \gets  \text{Construct Valid Sets (0)}\)
    \State \(i\gets 0\) \{AUV counter\}
    \State \(n\gets \) number of AUVs
    \While{\(i\not=n\)}
    \State \(\text{trajectory, success} \gets \text{Perform Planning}(i)\)
    \If {success}
    \State \(\text{trajectories}_i \gets \text{trajectory}\)
    \State \(\text{conflict} \gets \text{Construct Valid Sets}(i)\)
    \If {conflict}
    \State \(\text{Clear Valid Sets} (i+1)\)
    \State \(\text{Add Conflict}(i, \text{conflict})\)
    \Else
    \State \(\text{Clear Conflicts} (i+1)\)
    \State \(i \gets (i+1)\)
    \EndIf
    \Else
    \State \(i \gets (i-1)\)
    \If {\(i<0\)}
    \State \textbf{return} \(\emptyset\) \{Planning Failed\}
    \EndIf
    \EndIf
    \EndWhile
    \State \textbf{return} $\text{trajectories}$
    \EndProcedure
  \end{algorithmic}
\end{algorithm}

The overall planning framework begins by predetermining the order of planning.
The valid set for the first AUV is then immediately calculated and a path to the
AUV's target location is planned. Upon successful planning, the trajectory for
the first AUV is used to determine the valid sets for the second AUV and then
path planning is performed for the second AUV. This process then continues, with
the trajectories of the already planned AUVs used to determine the valid
sets of the next-to-be-planned AUV until planning has been performed for all
AUVs in the network.

If during the valid set construction phase any valid set is found to be entirely
empty, that is there is no valid location for the corresponding AUV and
timestep, the time and location of the previous AUV at that time is then
considered a `conflict' and the path of the preceding AUV must be replanned such
that it does not enter the conflict state. If planning for any AUV $i$ fails,
then the planner reverts to AUV $i-1$ and the process continues. These two
mechanisms, conflicts and replanning, allow for the planner to more flexibly
handle poor configurations and attempt to replan based on knowledge of failed
planning attempts without introducing any computational overhead in the case of
successful planning.

\begin{algorithm}
  \caption{Construct valid sets for AUV $i$ based on planning conflicts and
    the trajectories of the previous AUVs} \label{alg:auv-0-feas}
  \begin{algorithmic}[1]
    \Function{Construct Valid Sets}{$i$}
    \State $P_{i,0} \gets $ AUV $i$ start location
    \State $V_{i,0} \gets P_{i,0}$
    \State $t \gets 0$
    \State $x \gets $ goal location of AUV $i$
    \While{$x \not\in P_{i,t}$}
    \State $N\gets$ all neighbors of $V_{i,t}$
    \State $P_{i,t+1} \gets (V_{0,t} \cup N) \setminus \text{conflicts}$
    \If {$i=0$}
    \State $V_{i,t+1} \gets P_{i,t+1} $
    \ElsIf {$i=1$}
    \State $C_{i, t+1} \gets \text{Connected States}(t+1)$
    \State $V_{i,t+1} \gets P_{i,t+1} \cap C_{i, t+1}$
    \Else
    \State $R_{i, t+1} \gets \text{Rigid States}(t+1)$ \label{alg:check-rigid}
    \State $V_{i,t+1} \gets P_{i,t+1} \cap R_{i, t+1}$
    \EndIf
    \If {$V_{i,t+1} = \emptyset$}
    \State $\text{loc} \gets \text{Get AUV Location}(i, t+1)$
    \State $\text{conflict} \gets (\text{loc, t+1})$
    \State \textbf{return} $\text{conflict}$ \{Returning state as conflict\}
    \EndIf
    \State $t\gets t+1$
    \EndWhile\label{euclidendwhile}
    \State \textbf{return} $V_i$ \{Returning valid sets\}
    \EndFunction
  \end{algorithmic}
\end{algorithm}
\section{Results and Discussion}
\begin{table*}[ht]
  \centering
  \caption{Information and results from simulated experiments. Each test case represents a unique set of obstacles, goal locations, and number of AUVs in the network. We present the planning time required, the makespan of the trajectories, the average and maximum localization errors, and the percentage of the trajectory which the network was in a rigid configuration.
  }

  \begin{tabular}{| c | c | c | c | c | c | c | c |}
    \hline
    {\bf Test Case} & {\bf Algorithm} & {\bf \# of AUVs} & {\bf Planning Time (s) } & {\bf Makespan} & {\bf Avg. Localization Error} & {\bf Max. Localization Error} & {\bf \% Rigid } \\

    \hline
    \multirow{2}{*}{1}
    & { RCGP (\bf Ours) } & 8 & \textbf{8.911} & \textbf{33} & \textbf{0.948} & \textbf{3.067} & \textbf{100}  \\
    & { RRT }             & 8 & 9.958 & 75 & 1.782 & 9.255 & 53  \\

    \hline
    \multirow{2}{*}{2}
    & { RCGP (\bf Ours) } & 6 & \textbf{7.141} & \textbf{32} & \textbf{0.338} & \textbf{1.874} & \textbf{100}  \\
    & { RRT }             & 6 & 8.315 & 72 & 1.056 & 11.044 & 75  \\

    \hline
    \multirow{2}{*}{3}
    & { RCGP (\bf Ours) } & 8 & 7.513 & \textbf{27} & 1.033 & \textbf{3.124} & \textbf{100}  \\
    & { RRT }             & 8 & \textbf{0.389} & 43 & \textbf{0.915} & 4.065 & 42  \\

    \hline
    \multirow{2}{*}{4}
    & { RCGP (\bf Ours) } & 6 & 3.154 & \textbf{28} & \textbf{0.197} & \textbf{0.581} & \textbf{100}  \\
    & { RRT }             & 6 & \textbf{0.354} & 46 & 0.392 & 1.445 & 91  \\

    \hline
    \multirow{2}{*}{5}
    & { RCGP (\bf Ours) } & 20 & 163.0 & \textbf{22} & 1.176 & \textbf{2.311} & \textbf{100}  \\
    & { RRT }             & 20 & \textbf{0.598} & 41 & \textbf{1.037} & 2.922 & \textbf{100}  \\

    \hline
  \end{tabular}
  \label{tab:planning-results}
\vspace{0.3cm}
\end{table*}
We tested our rigidity-constrained graph planning (RCGP) framework over a number
of two-dimensional simulated environments with varying numbers of AUVs and
obstacles. One of the tested environments and planning objectives is shown
Figure~\ref{fig:planning-problem} for reference.

\subsection{Implementation}
We compare statistics on timing, planning, localization, and rigidity for our
planning technique to a priority planning version of the rapidly-exploring
random tree (RRT) algorithm in which the only planning constraint was that no
two robots could occupy the same location at the same time. We also tested a
fully coupled probabilistic roadmap planner but found that it was unable to
complete planning for a small number of robots after ten minutes so we do not
show the results. Simulations were programmed in Python and were run on an Intel
i5-8300H processor. Results are shown in Table~\ref{tab:planning-results}.

\begin{figure}[!t]
\begin{tabular}{cc}
\hspace{-5mm}
\subfloat[Planning problem: dots are start positions and crosses
are goals] {
\includegraphics[trim=0mm 0mm 0mm 0mm, clip, height=0.6\columnwidth,
width=\columnwidth]{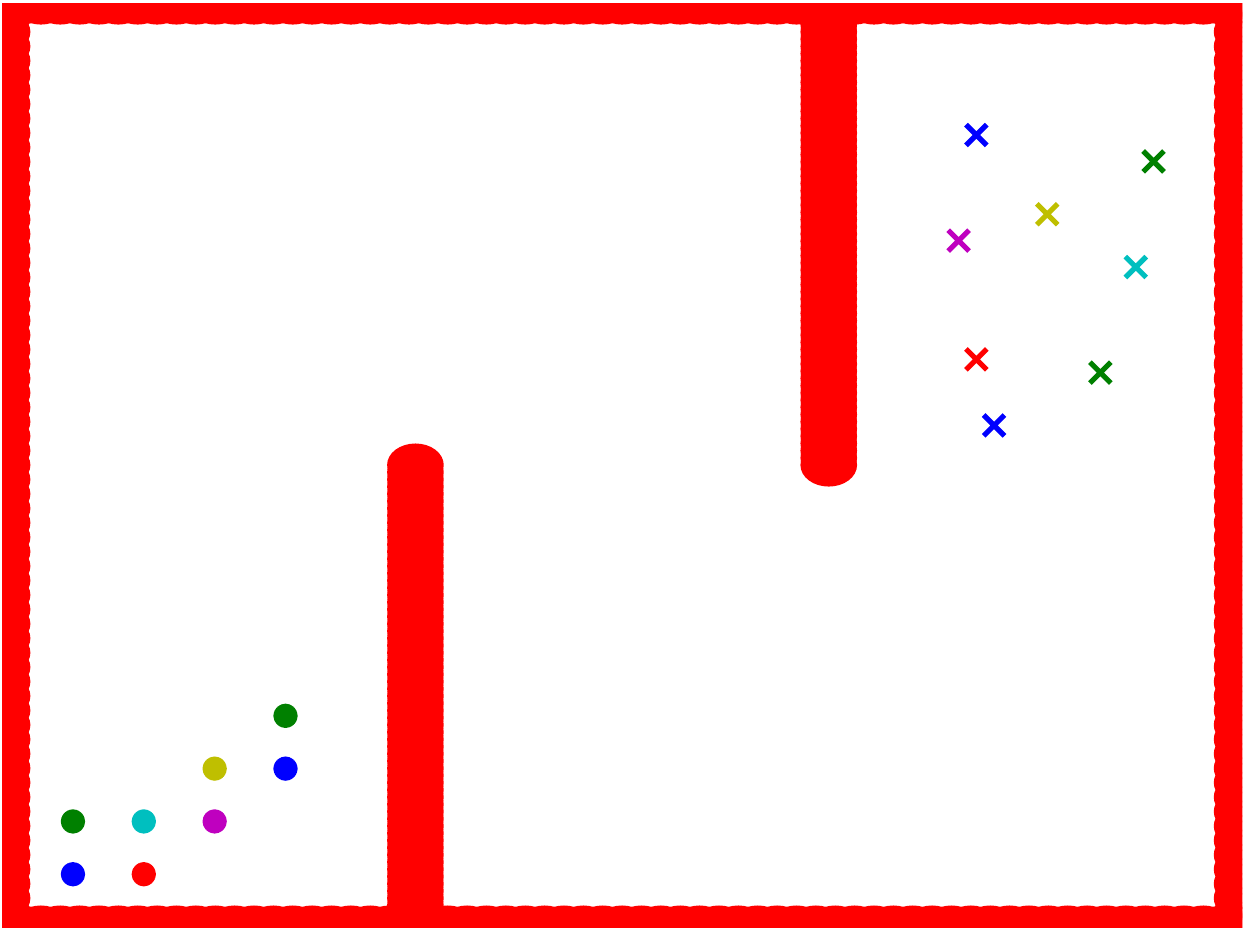} \label{fig:planning-problem}}
\\ \\
\hspace{-5mm}
\subfloat[Planning graph for the planning problem shown above]{
\includegraphics[trim=0mm 0mm 0mm 0mm, clip, width=0.999\columnwidth]{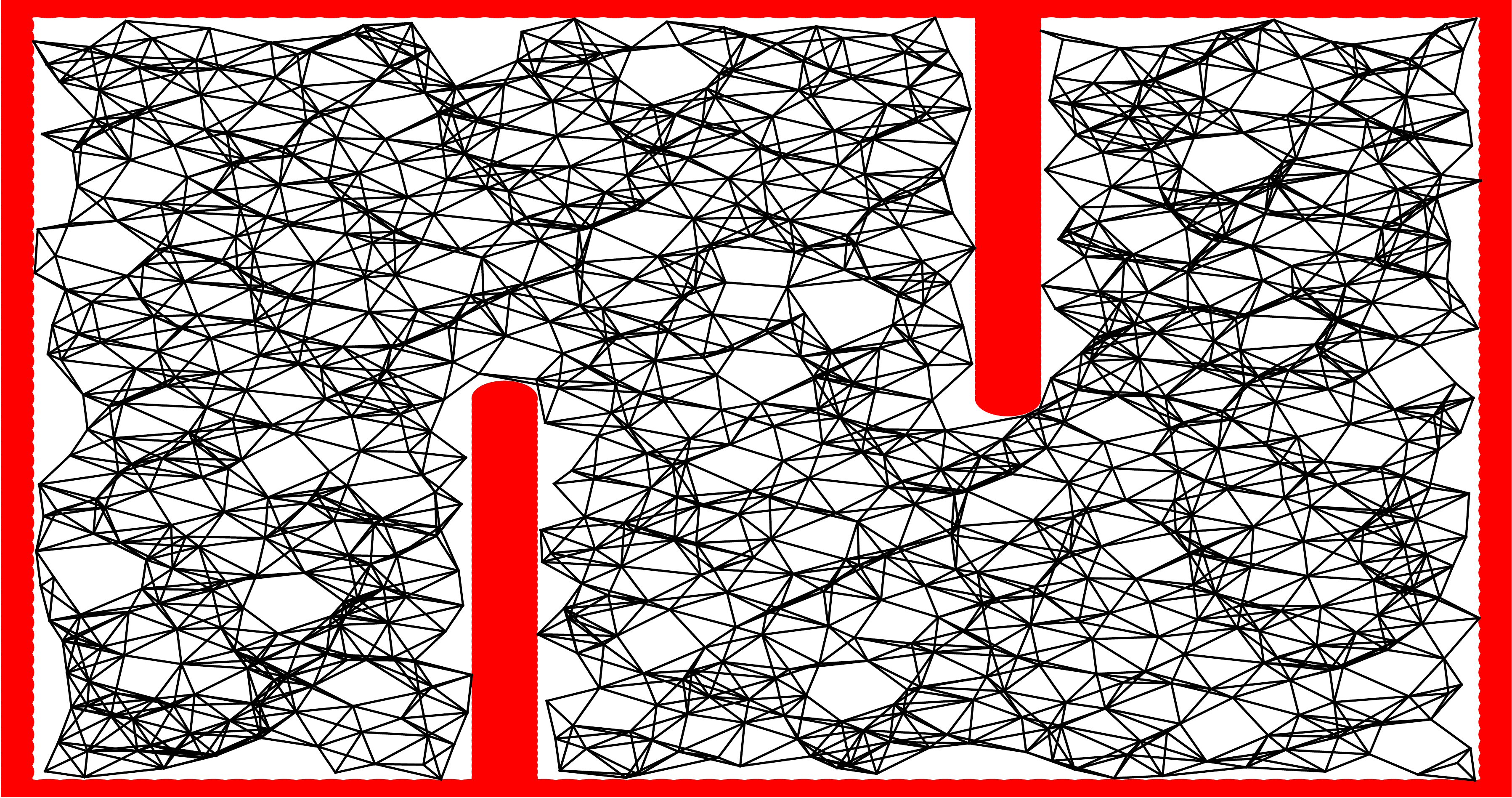}\label{fig:planning-graph}}
\end{tabular}
\caption{Example planning problem and corresponding planning graph. The red outlines denote obstacles to avoid. Colors are used to show correspondence between the start and goal positions for each simulated AUV}
\end{figure}

To generate our planning graph we uniformly sampled the space over the
environment at predetermined intervals and made edges between every sampled node
and its nearest neighbors within a distance of 2 units. Such a planning graph
can be seen in Figure~\ref{fig:planning-graph}. We removed any edges that
intersected obstacles in the environment. We assumed each robot to be a point
robot and that there were no kinematic restrictions on a robot's movement. To
then perform planning on this graph we used a heuristic directed A* search with
the Euclidean distance from a node to the goal location as the heuristic cost.
We chose the minimum allowable rigidity to be 0.1, as this qualitatively
appeared to provide a balance between localization quality and the consistent
ability to successfully find valid trajectories.

\subsection{Localization}
To compare the ability to localize over the course of a given path we
implemented the sensor network localization approach of \cite{Biswas2006} and
report both the mean error and the max error of the localization results for a
given trajectory, where error is calculated as the Euclidean distance between
the SNL localization results and the ground truth locations. To allow for
absolute localization, as opposed to just relative, at each timestep three AUVs
were randomly selected as having a known location for the purpose of SNL.

As expected, the localization results for trajectories generated by our
algorithm generally had similar or reduced average localization error than the
trajectories generated by the RRT algorithm. Notably, the maximum localization
error for RCGP trajectories was lower in every test case, and often was a
substantial amount below the maximum localization error for RRT trajectories.
This would support the claim that the use of rigidity-constrained planning
improves ability to perform range-only localization.

In addition, while percent time rigid is useful metric, it should be noted that
this statistic still misses many important details, as there are many different
qualities of non-rigidity and network configuration which all affect
localizability to varying degrees. This helps explain why there is no strong
relationship between percent time rigid and localization error beyond the fact
that trajectories which have any amount nonrigid time appear to have increased
localization errors. Additionally, it is important to note that other
localization algorithms will report different localization results, though the
general relationship between error and rigidity is expected to hold true.

\subsection{Planning}
To evaluate the feasibility of each planner we report the makespan and time
required to plan the full set of trajectories. The makespan is the time elapsed
from the start of the trajectories to when the last AUV reaches its goal
position. In the case of RGCP the planning time includes the time required to
build the planning graph.

In environments with marked complexity or more challenging obstacles, notably
Test Cases 1 and 2, it was found that the planning time was comparable between
the two approaches and. However, in more simple environments with randomly
placed obstacles and a large amount of free space the RRT was able to plan
trajectories an order of magnitude faster with a moderately sized number of AUVs
(Test Cases 3 and 4) and three orders of magnitude faster with a larger number
of AUVs (Test Case 5). This reveals a trend of the RRT planner generally scaling
linearly with the number of AUVs while the runtime of our planner does not scale
as well. Profiling of our planner reveals that over ninety-nine percent of the
computation of our planner is due to checking the rigidity of a location, as
shown in step \ref{alg:check-rigid} of Algorithm~\ref{alg:auv-0-feas}.

Beyond planning time, we note that our RGCP algorithm finds trajectories which
generally have a much reduced makespan compared to our RRT implementation. This
difference between our RRT implementation and our RGCP algorithm is largely due
to the use of A* to perform planning on the graph. This incentives the RGCP
planner to minimize travel time. In comparison, the RRT implementation used
would often generate inefficient trajectories with no weight given to the travel
time. We do not compare the resulting makespans to the theoretically optimal
makespan, but these results along with visual inspection of the trajectories
indicate that the RGCP planner does generate time-efficient trajectories.

\section{Conclusions}
\comment{
TODO could add something about planning considering which robots are "anchors"
TODO discuss the results a little more
}

We were successfully able to construct a rigidity-constrained planning framework
which was able to reliably plan efficient, rigidity-constrained trajectories.
The results seen in Table~\ref{tab:planning-results} show that the planner is
capable of planning paths for a moderate number of AUVs in a feasible amount of
time while meeting a minimum network rigidity at every timestep. The use of our
network rigidity measure is supported by the given localization results, which
show higher localization accuracy for our planner than our RRT implementation,
which does not guarantee a minimum network rigidity.

Future work in this field of rigidity-constrained planning should explore the
use of precomputed formations to avoid the need for evaluating a given network's
rigidity at planning-time, as the majority of time spent in planning is on
evaluating the rigidity. Furthermore, the presented priority-planning approach
has two distinct drawbacks; the presented framework can eliminate valid
trajectories from start to goal configurations and the quality and success of
the planning is highly dependent on priority ordering. Future work to address
these issues could consider other representations of the rigidity-constrained
planning problem which either does not use priority-based planning or does not
impose such strict planning constraints as rigidity at every step of the
planning sequence.
\section{Acknowledgements}

This work was partially supported by ONR grant N00014-18-1-2832,
ONR MURI grant N00014-19-1-2571 and the MIT-Portugal program.
\balance
\printbibliography
\clearpage
\newcounter{y}
\setcounter{y}{15}

\begin{figure*}[]

    \begin{tabular}{cc}

        \subfloat[Starting Position] {
            \includegraphics[trim=0mm \value{y}mm 0mm \value{y}mm, clip, width=0.4\paperwidth]{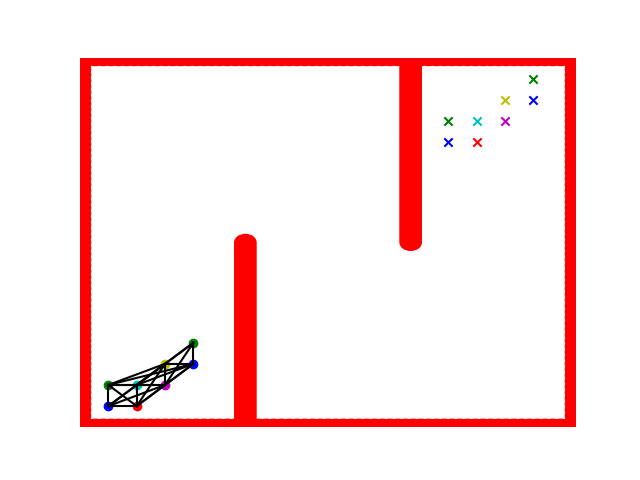}}

         &

        \subfloat[Positions at timestep 5]
        {\includegraphics[trim=0mm \value{y}mm 0mm \value{y}mm, clip,
                width=0.4\paperwidth]{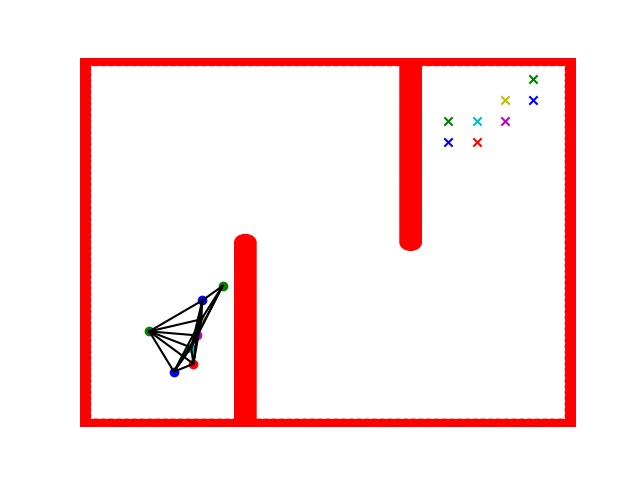}\label{fig:rigid-traj-1}
        }
        \\

        \subfloat[Positions at timestep 10]
        {\includegraphics[trim=0mm \value{y}mm 0mm \value{y}mm, clip,
                width=0.4\paperwidth]{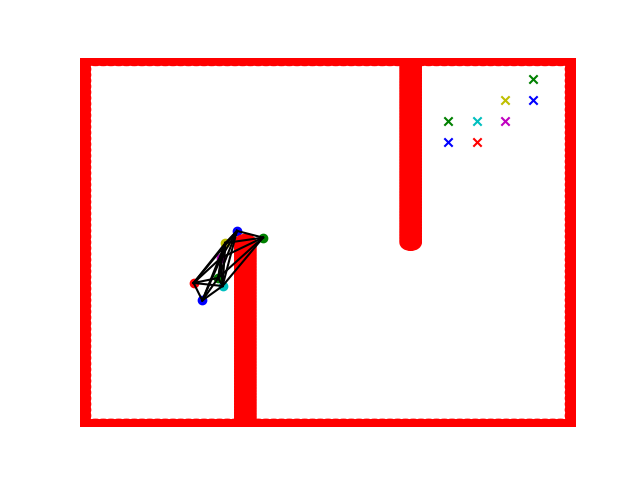}}

         &

        \subfloat[Positions at timestep 15]
        {\includegraphics[trim=0mm \value{y}mm 0mm \value{y}mm, clip,
                width=0.4\paperwidth]{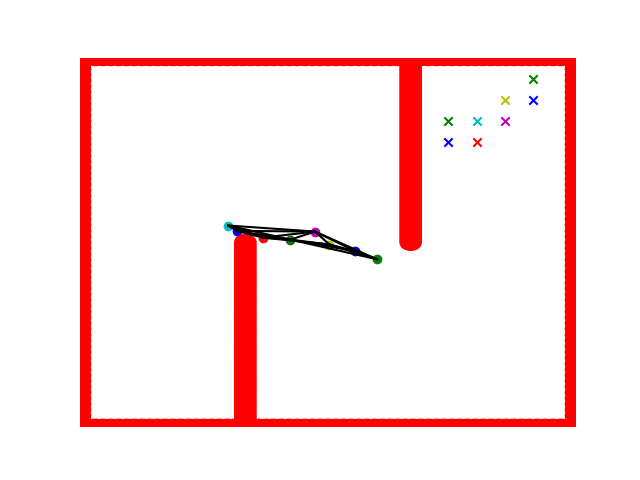}\label{fig:rigid-traj-2}
        }
        \\

        \subfloat[Positions at timestep 20]
        {\includegraphics[trim=0mm \value{y}mm 0mm \value{y}mm, clip,
                width=0.4\paperwidth]{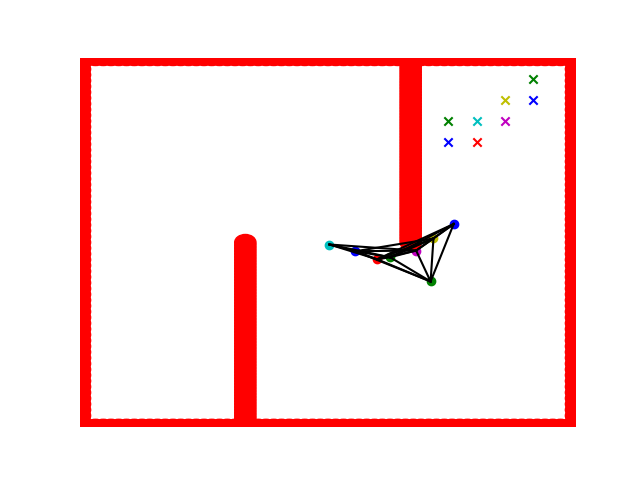}\label{fig:rigid-traj-3}
        }

         &

        \subfloat[Positions at timestep 25]
        {\includegraphics[trim=0mm \value{y}mm 0mm \value{y}mm, clip,
                width=0.4\paperwidth]{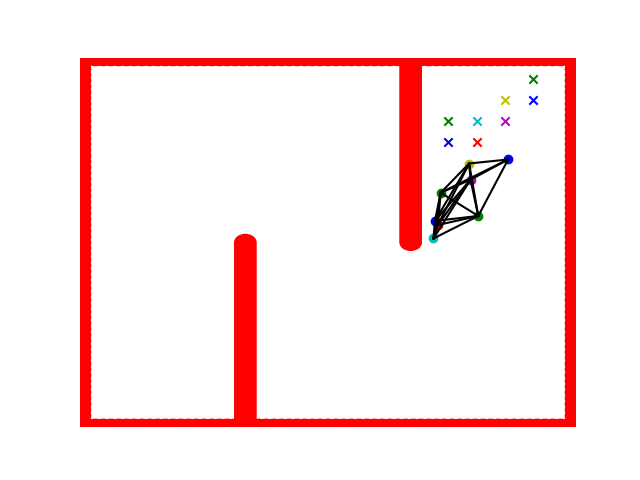}}
        \\

        \subfloat[Positions at timestep 30]
        {\includegraphics[trim=0mm \value{y}mm 0mm \value{y}mm, clip,
                width=0.4\paperwidth]{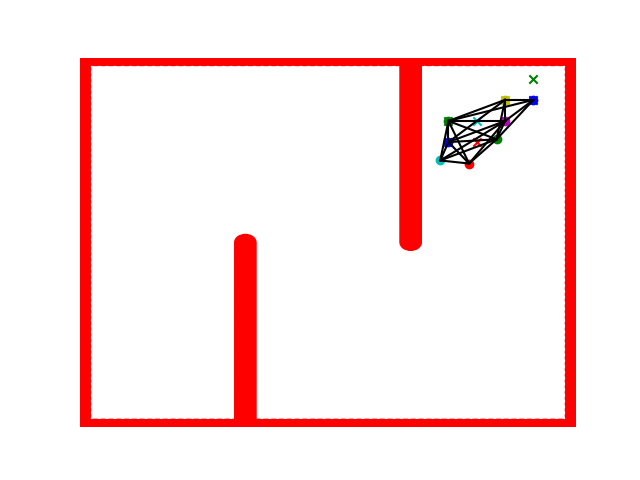}}

         &

        \subfloat[Arrived at goal positions]
        {\includegraphics[trim=0mm \value{y}mm 0mm \value{y}mm, clip,
                width=0.4\paperwidth]{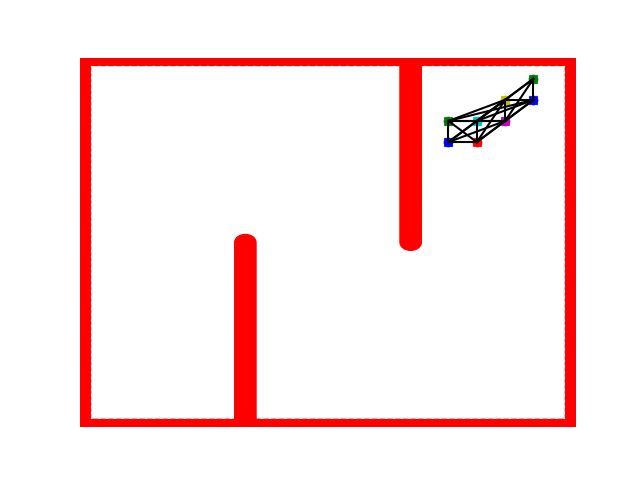}}
    \end{tabular}

    \centering
    \caption{Snapshots of a planned trajectory. It can be notices that in sections where a naive planner might collapse into a low-rigidity configuration (such as rounding a corner) the rigidity-constrained planner will have a small number of robots which provide triangulation for all of the others. This can be observed specifically in \Cref{fig:rigid-traj-1},  \Cref{fig:rigid-traj-2}, and \Cref{fig:rigid-traj-3}}



\end{figure*}
\end{document}